\documentclass[letterpaper, 11pt]{article}

\usepackage{naaclhlt2010}
\usepackage{latexsym}
\usepackage{amssymb}
\usepackage{times}
\usepackage[margin=1in]{geometry}
\usepackage{amsmath,amsfonts,amsthm}
\usepackage{graphicx}
\usepackage{algorithm}
\usepackage[noend]{algorithmic}
\usepackage{cite}
\usepackage[numbers]{natbib}

\title{Comparison of SVM Optimization Techniques in the Primal}
\author{Diane Duros \\
	Johns Hopkins University\\
	dduros1@gmail.com
\And Jonathan Katzman\\
	Johns Hopkins University\\
	jonathan.d.katzman@gmail.com}

\begin{document}
\maketitle

\begin{abstract}
This paper examines the efficacy of different optimization techniques in a primal formulation of a support vector machine (SVM).  Three main techniques are compared.  The dataset used to compare all three techniques was the Sentiment Analysis on Movie Reviews dataset, from kaggle.com.
\end{abstract}
                                                                                                                                                                                                                                                                           
\section{Introduction}
Most SVM literature states the primal optimization, then proceeds to the dual formulation without providing significant detail on training an SVM using the primal optimization problem.  Learning an SVM is typically viewed as a constrained quadratic programming problem.

Our goal is to analyze three different primal optimization methods in the context of a large, text-based dataset.  Given a training set $\{ (x_i, y_i) \}_{1\le i\le n}, x_i \in \mathbb{R}, y_i \in \{1, -1\} $, the primal optimization problem is:
\begin{gather} min_{w,b} \mid \mid w \mid \mid ^2 + C\sum_{i=1}^n \xi_i^P\notag\\ 
\text{under constraints } y_i(w\cdot x_i + b) > 1; \xi_i, \xi_i > 0\label{primalobj}\end{gather}

We use a hard-margin SVM, where $C=0$ - that is, we would like every data point to be outside of the margin around the hyperplane.  The values $\xi_i$ allow for some slack, but by setting $C$ to $0$, we remove this possibility.

The optimization algorithms we consider are gradient descent, Newton's method, and the Pegasos algorithm, which is an application of a stochastic sub gradient method.  Other algorithms that are related to these are the NORMA algorithm\cite{norma}, which is another application of stochastic gradient descent, and SGD-QN \cite{sgdqn}, which combines stochastic gradient descent with a quasi-Newton method.  Other previous work in Newton's method \cite{chapelle2007training} has used the USPS dataset, which is significantly smaller than our data set.

\section{Background}

\subsection{Data}
We retrieved our data from the problem ``Sentiment Analysis on Movie Reviews," from kaggle.com.  The data originates from the Rotten Tomatoes dataset and consists of phrases that have been assigned sentiment labels, where the sentiment labels are \{ 0: negative, 1: somewhat negative, 2: neutral, 3: somewhat positive, 4: positive \}.\\
\indent Initially, we collapsed the labels to binary labels, where an original label of 3 or 4 was considered positive, and 2 or below became a negative label.  Later, we created a MulticlassSVM to handle the non-binary labels.  \\
\indent In addition to collapsing the labels, we also had a number of options for processing our data.  We ignored punctuation such as  ``,", determining that it was unlikely to contribute significantly to the sentiment while appearing often in both positive and negative phrases.  We also ignored upper/lower case distinctions, which collapsed our corpus from $\sim$18000 words to $\sim$16000 words.\\
\indent When we compare our data to that of the USPS dataset, USPS has 7291 data instances with 256 features, so our data is at least an order of magnitude larger than this.
\begin{table}
\centering
\label{stats}
\begin{tabular}{cc}
\textbf{Data Statistics}&\\
\hline
Number of data instances & 150606\\
Number of distinct words & 18226\\
Avg. freq. of words per phrase & 6.85\\
Avg. freq. of phrases per word & 55.21
\end{tabular}
\caption{Data statistics on Sentiment Analysis on Movie Reviews dataset}
\end{table}

\subsection{Features}
We generated features based on a bag of words model, where a phrase is represented as the multiset of its words.  There are two options: binary features (where the feature indicates the existence of a word in a phrase) and continuous features (where the feature indicates the frequency of a word in a phrase).

We have as many features for each instance as there are words in the corpus (18226), but each feature vector is sparse, as we can see in table \ref{stats}.  Since there's an average of 6.85 words per phrase (data instance), the other 18220 elements of the feature vector will be 0.  We did not implement feature selection to only utilize the most significant features, as we believe that feature selection would not be able to maintain a balance between reducing the number of features and maintaining information for each phrase (if we removed too many words, many phrases would have most, or maybe all, of their words removed, becoming useless).

\section{SVMs}
We chose to approach this data set with SVMs, because they provide a framework for various optimization methods, as well as generalizing to multi-class data.  This also allowed us to build upon work that we'd done in class, using the gradient descent method as a baseline to compare with our other optimization methods.

SVMs are binary linear classifiers.  In training, they create a hyperplane between two classes of data, where the hyperplane maximizes the margin between the two classes.  The main assumption that we have made about our data is that it is linearly separable.  Since our data is text-based, one example that could cause non-linear separability is sarcasm: suppose a reviewer uses a ``positive" word sarcastically.

Our optimizers return the coefficients of the hyperplane that maximizes the margin, where we have as many coefficients as we do features.  The general objective function is of the form $f(w) = L(w) + r(w) $, where $L(w)$ is a convex measure of loss and $r(w)$ is a convex regularization, so our optimization goal is to minimize the loss function.

\textbf{Basis} To calculate a basis, we take all data points that constitute support vectors (that is, whose margins are less than one), and calculate their average distance from the hyperplane described by the weights $w$.  We use the average distance, because it is a more robust measure for implementation; if we summed these values, we would overfit the data. %Taking just the sum of their distances would more accurately optimize the SVM equation, but can overfit the data - the average distance is a more robust measure. 

\subsection{Multi-class SVM}
In order to classify data into 5 classes, we needed to expand our SVM to handle more than just binary labels.  Since two-class problems are easier to solve, we will generate multiple pairwise SVMs for multi-class classification, using the ``one against one" method \cite{milgram2006one}.

To enforce ordinal ranking ($0 < 1 < 2 < 3 < 4$), instead of generating pairwise SVMs between all pairs, we only created pairwise SVMs for label pairs \{ (0,1), (1,2), (2,3), (3,4) \}.  This way, we never attempt to classify between classes that aren't directly related via the ranking inequality.

Once each pairwise SVM is trained, according to a user defined optimization method, there is an additional training step.  For prediction in the binary SVM, we compute $E(W)$, where $W$ represents the weights trained by the SVM.  If $E(W) \ge 0$, we classify the example as positive, otherwise, it is negative.  For the multiclass SVM, we compute E(W) for each pairwise SVM, and store the values in $S$.  Then, we compute $P(label \mid S) = \frac{P(label, S)}{P(S)}$, that is, we count the number of times the true label occurs with the array $S$.  Then, when we predict an instance, we calculate $S$, then determine which label is most likely (has the highest probability).

This introduces the problem that, if we have never seen the true label and $S$ together, we will never be able to predict it.  In addition, if we train on a restricted dataset, it is possible that we will not see all possible values of S.  In this case, if we encounter a value of S that we didn't see in training, we classify based on the overall probability of the labels, given the training data (the probability is the proportion of times we've seen each label in training data to the number of all training instances.).

For this multi-class SVM, we need to achieve accuracy above 0.2 (1/5) to improve upon random chance, whereas we need to have accuracy greater than 0.5 for the binary SVM.

\section{Primal Optimization Methods}
Many SVM packages optimize the dual form of the SVM, but we chose to explore different methods for optimizing the primal problem.

For a linear SVM, both the primal and the dual are convex quadratic programs, so an exact solution exists.  Chapelle shows not only do  the primal and dual optimization methods reach the same result, but that when an approximate solution is desired, primal optimization is superior\cite{chapelle2007training}.  The  dual program solves for a vector that is as long as the number of training instances, and the primal program solves for a vector that is as long as the number of features.  As seen in table \ref{stats}, we have 150,000 instances, but only 18,000 features, so the primal problem will be more efficient to solve for our data set.

We examined three different optimization methods: gradient descent, Newton's approximation, and stochastic subgradient (where we used the Pegasos algorithm).  Since the hinge loss is non-differentiable, Chapelle considered smooth loss functions instead of the hinge loss\cite{chapelle2007training}, while the Pegasos algorithm\cite{pegasos} uses sub-gradients.

\subsection{Gradient Descent}

We use gradient descent as a baseline with which to compare our two more complex methods.
Gradient descent is a classical optimization technique, since the gradient of a function points in the direction of greatest increase, and therefore the negative gradient points in the direction of greatest decrease.  It uses the update step:
\begin{equation}
	w' = w - \eta\bigtriangledown L(x,y; w)
\end{equation} 
where $\eta$ is the learning rate, $L(x,y; w)$ is the loss function for a data point given the current hyperplane coefficients w.  We used a default step size (learning rate) of .001.  We sacrificed runtime and  used a small step size so that we wouldn't overshoot and miss the minimum.  In this method, we use the entire training set to compute the gradient.\cite{menon}

\textbf{Loss function} The loss function used is quadratic loss ( which is differentiable everywhere unlike the hinge loss), the $L_2$ penalization of training errors, $$ L(y_i, f(x_i)) = max (0, 1-y_if(x_i))^2 $$

\textbf{Runtime} Gradient descent is linear in the number of training instances, iterations, and number of features, which in this case is quite large.

\subsection{Newton's Approximation}

As seen in \citeauthor{chapelle2007training}, we can write \eqref{primalobj} as an unconstrained optimization function:
\begin{equation}
	\lambda \beta^TK\beta + \sum_{i=1}^n L(y_i, K_i^T\beta)
	\label{newton}
\end{equation}
 where $\lambda$ is a regularization parameter, L is the loss function, and $K_i$ is the ith column of a kernel K.

 \textbf{Loss Function}  The loss function used is quadratic loss, the $L_2$ penalization of training errors, $$ L(y_i, f(x_i)) = max (0, 1-y_if(x_i))^2 $$  If the loss on a point $x_i$ is nonzero, then $x_i$ is a support vector.  The gradient of \eqref{newton} with respect to $\beta$ is $$ \bigtriangledown = 2(\lambda K \beta + K I^0(K\beta -Y) ) $$ and the Hession is $$ H = 2(\lambda K + K^0 K) $$  We can combine these to see that after the Newton update step\footnote{This assumes that K is invertible.}, \begin{equation} \beta = (\lambda I_n + I^0K)^{-1}I^0Y  \end{equation}  If $K_{sv}$ is the sub matrix corresponding to the support vectors, then since $\lambda I_n + I^0 K = 0$, the final update is \begin{equation}  \beta = \begin{matrix} (\lambda I_{n_{sv}} + K_{sv})^-1 Y_{sv} \\ 0 \end{matrix}\end{equation}

When we attempted to run our implementation of this method on our full training set, we realized that it would not terminate in a reasonable amount of time for this project.  Since the algorithm initially restricts data to 1000 samples, then recursively trains on double the data until finally training on the full dataset, there is a cubic increase in time if the number of support vectors increases with size of the dataset, which it does.  On our data, it would take 9 recursive calls to fully train the model.

\begin{table}
\centering
\begin{tabular}{c|cc}
\textbf{Recursion step} & \textbf{number of instances} & \textbf{time} (sec)\\
\hline
1 & 1000 & 24\\
2 & 2000 & 107\\
3 & 4000 & 410\\
4 & 8000 & 1665\\
5 & 16000 & 6805\\
\end{tabular}
\caption{Timing data for the recursive step of Newton's Approximation}
\end{table}

Since we could not train this model on our full dataset, we created a restricted dataset of 1000 instances for evaluation purposes.  We also restricted the number of iterations to 5, since Chapelle claims that the algorithm should converge to the solution within 5 iterations.

This method approximates the inverse Hessian, and uses it to scale the gradient at each iteration.  Methods like this tend to be avoided for large-scale, batch problems, because, as we have discovered, this method does not scale well to large data.

\textbf{Runtime}  The runtime depends on the complexity of one Newton step, $O(nn_{sv} + n^3_{sv})$, since we converge in a constant number of iterations.  

\begin{algorithm}
\caption{Newton's Approximation}
\begin{algorithmic}
\STATE function $\leftarrow$ primalsvm(Y, $\lambda$)
\STATE n $\leftarrow$ length(X)
\IF {$n > 1000$}
\STATE $n_2 \leftarrow n/2$
\STATE $\beta \leftarrow$ primalsvm($Y_{1\ldots n_2}$, $\lambda$)
\STATE sv $\leftarrow$ non-zero components of $\beta$
\ELSE
\STATE sv $\leftarrow \{ 1, \ldots, n \}$
\ENDIF
\REPEAT
\STATE $\beta_{sv} \leftarrow (K_{sv} + \lambda I_{n_{xv}})^{-1}Y_{sv}$
\STATE Other components of $\beta \leftarrow 0$
\STATE sv $\leftarrow$ indices $i$ such that $y_i[K\beta]_i < 1$
\UNTIL {sv has not changed}
\end{algorithmic}
\end{algorithm}
\textbf{RBF Kernel}  We used a radial basis function (RBF) kernel for this method, because that is the original kernel used by Chapelle.  $$ K(x, x') = exp(-\frac{\mid \mid x-x' \mid \mid^2_2 }{2\sigma^2}) $$After getting poor performance from this method, we further examined our kernel, and determined that we should not have used an RBF kernel for this problem. The RBF kernel acts as a Gaussian filter, which are often used for smoothing images, so the RBF kernel is a filter that selects smooth solutions.\footnote{http://charlesmartin14.wordpress.com/2012/02/06/kernels\_part\_1/ }  This doesn't make sense in the context of text data, because you can't view the data in terms of signals with a frequency domain.  This kernel was logical for the USPS dataset Chapelle used, because it is an image dataset.  A better kernel for us would have been the linear kernel.

\subsection{Stochastic Subgradient}
Stochastic gradient descent is significantly faster than methods like gradient descent that use the true gradient at each iteration.  Instead, it approximates the gradient on fewer training examples.  The stochastic sub gradient method we implemented is the Pegasos algorithm by \citeauthor{pegasos}, which is a stochastic gradient descent method that also has a projection step.  The objective to be minimized is \begin{equation} min_w \frac{\lambda}{2} \mid \mid w \mid \mid ^2 +\frac{1}{m} \sum_{(x, y) \in X} L(w; (x,y)) \end{equation}

At each iteration, we select a random subset of training examples, and update the weight vector with the subgradient of the objective function evaluated on the subset.  Then we project the vector onto a sphere with radius $1/\sqrt{\lambda}$, because the optimal weight vector must lie in this ball (see \citeauthor{menon}), due to the strong duality theorem.

\begin{algorithm}
\caption{Pegasos Stochastic Subgradient Method}
\begin{algorithmic}
	\FOR {$t = 1\ldots T$}
		\STATE Pick random $A_t \subset T$ s.t. $\mid A_t \mid = k$
		\STATE$ M \leftarrow \{ (x,y) \in A_t: 1-y(w\cdot x) >0 \}$
		\STATE $\bigtriangledown_t \leftarrow \lambda w_t - \frac{1}{\mid M \mid} \sum_{(x,y) \in M} yx$
		\STATE Update $w_{t+\frac{1}{2}} \leftarrow w_t - \frac{1}{\lambda t}\cdot\bigtriangledown_t$
		\STATE Let $w_{t+1} \leftarrow min \left( 1, \frac{1}{\sqrt{\lambda} \mid \mid w_{t+1/2} \mid \mid} \right) w_{t+\frac{1}{2}}$
	\ENDFOR
	\RETURN $w_{T+1}$
\end{algorithmic}
\end{algorithm}

\textbf{Loss Function} The Pegasos algorithm uses the hinge loss function $$ L(w; x,y) = max\{0, 1-y\langle x, y\rangle\} $$

\textbf{Runtime} The runtime is independent of the number of training examples, and the algorithm finds an $\epsilon$-accurate solution in $O(\frac{d}{\lambda \epsilon})$ time, where d is the number of non-zero features in each training example.

\section{Code}
We created an SVM framework in Python, as well as a custom DataParser.  Our SVM class takes an Optimizer object as input, where the Optimizer calculates weights for the SVM according to the optimization function encoded in the object.   We did not use any SVM packages, and only used the numpy package for evaluation purposes.

\subsection{Evaluation}
For evaluation, we created a CrossValidationTester, so that we could train and test on the training dataset from kaggle.com.  For each round, the CrossValidationTester randomly orders the data, then selects 10\% to be left out for testing purposes, while the rest is used for training the model; we used a default of 10 rounds, then averaged the accuracy and timing results across rounds.  The only model that we didn't use this for was the Newton approximation, because it took too long to run.

\section{Results}

\subsection{Cross Validation Results}

To find the best accuracy for each algorithm, we tested different parameter settings.  For gradient descent, we iterated over the set of learning rates \{ .01, .02, .03, .04, .05, .1, .2, .3, .4, .5, 1, 2, 3, 4, 5  \} to find the optimal learning rate, then used this value to find the number of iterations needed for convergence.  Similarly, for the Pegasos method, we iterated over the set of ball diameters \{ .001, .01, .1, 1, 10 \} to find the optimal $\lambda$, then used this to determine the best sample size and number of iterations.  The optimal parameters for each algorithm under each data representation setting are available in the results file we're providing with our project.

For each parameter setting, we ran 10-fold cross validation to determine the average accuracy and runtime.  Once we determined the "best" parameters for each model, we ran 10-fold cross validation on the training data under various data representation settings.  The modes we evaluated were binary (bin) versus multiclass (multi) SVMs, and the data settings were binary feature representation (bin) versus the frequency representation (freq).  Table \ref{table:fullset} shows the comparison between different settings for gradient descent (GD) and Pegasos stochastic subgradient (SSG) on the full data set, while table \ref{table:smallset} shows the comparison between settings for all three algorithms, where Newton's method is abbreviated as NM.

We were unable to evaluate Newton's on training sets larger than 1000 instances, but wanted to compare its results with the other two methods, so we used a limited data set to produce the results in table \ref{table:smallset}.

\begin{table}

\centering
\begin{tabular}{c|cccc}
\textbf{Alg.} & \textbf{Mode} & \textbf{Data} & \textbf{Accuracy} & \textbf{Time (sec)}\\
\hline
GD & bin & bin & .7328 & 184 \\
GD & multi & bin & .5119 & 300 \\
SSG & bin & bin & .7319 & 140 \\
SSG & bin & freq & .7312 & 67 \\
SSG & multi & bin & .5142 & 120 \\
SSG & multi & freq & .5141 & 122 \\
\end{tabular}
\caption{Results from running algorithms on full training set using cross validation.  We were unable to evaluate the full data set using Newton's approximation}\label{table:fullset}.
\end{table}

\begin{table}
\centering
\begin{tabular}{c|cccc}
\textbf{Alg.} & \textbf{Mode} & \textbf{Data} & \textbf{Accuracy} & \textbf{Time (sec)}\\
\hline
GD & bin & bin &  .8343 & 1\\
GD & bin & freq & .8404 & 1\\
GD & multi & bin & .6979 & 2\\
GD & multi & freq & .6929 & 3\\
SSG & bin & bin & .8485 & 2\\
SSG & bin & freq & .8485 & 3\\
SSG & multi & bin & .7061 & 1\\
SSG & multi & freq & .7172 & 1\\
NM & bin & bin & .7959 & 470 \\
NM & multi & bin & .6828 & 537 \\
\end{tabular}
\caption{Results from running algorithms on small (1000 data instances) cross validation set}\label{table:smallset}
\end{table}

On both datasets, the Pegasos method's results are nearly equal to those of gradient descent, but the runtime is much quicker.  On the small dataset, Newton's method performs almost equally well with the multiclass SVM, but is outperformed by both gradient descent and Pegasos on the binary SVM.  We can clearly see that this method took significantly longer, without yielding superior results.  Overall, we consider the Pegasos algorithm to be the most successful on this dataset.

\subsection{Comparison to Proposal}
We revised our proposal from focusing on solving the kaggle.com problem to using the dataset to explore different optimization techniques.  Our achievements did still align somewhat with what we initially planned to do, which was to first create a binary SVM classifier, then extend it to a multi-class SVM.  We created both successfully.  However, instead of exploring various feature representations of our data, we explored different primal optimization techniques.  Our goal was to implement three different algorithms, and we met this goal.  We did not generate results to submit to kaggle.com, preferring to evaluate our methods using 10-fold cross validation.

\nocite{*}
\bibliographystyle{plainnat} 
\bibliography{writeup}

\end{document}